\documentclass{article}
\usepackage{spconf,amsmath, amssymb,graphicx,hyperref,bm,multirow,booktabs,arydshln,courier,caption, enumitem, url}
\usepackage[table]{xcolor}
\hypersetup{hidelinks}
\definecolor{penguingreen}{RGB}{0,128,0}
\definecolor{penguinred}{RGB}{200,0,0}

\title{Learning Where to Look: A Shared Relative-Alignment Module for \\ Time-Series Forecasting and PPG-to-Vital-Sign Reconstruction}
\name{Ragamayi Puli$^{*}$, Shunya Nagashima$^{*}$\thanks{$^{*}$Equal contribution.}}

\address{Neurogica Inc.}
\begin{document}
\ninept
\maketitle
\begin{abstract}
PPG-to-vital-sign reconstruction turns a wrist-worn photoplethysmogram into clinical waveforms such as the ECG. Long-horizon multivariate time-series forecasting underpins planning in energy, weather, and traffic. Both generate a target sequence from a condition sequence, and current models hard-code where each target position reads it, as a same-position copy or seasonal recurrence, so neither transfers between tasks. We propose ROOSTER, one conditioning module that handles vital-sign reconstruction and time-series forecasting alike by learning this correspondence. Its core is a periodic-comb bias over the target--condition offset whose center, period, and sharpness are learned per head, so one module settles on the identity alignment or a seasonal lag and reports which it found. On vital-sign reconstruction from PPG, ROOSTER outperformed the published baselines on four heart-rate and respiratory-rate benchmarks. On multivariate time-series forecasting, it achieved the best horizon-averaged MSE on four benchmarks and outperformed the forecasting model it extends on 20 of 24 dataset--horizon settings under matched three-seed training. An ablation study indicated that the relative bias, not content matching, carried the alignment.
\end{abstract}
\newcommand{\parhead}[1]{\vspace{0.35\baselineskip}\noindent\textbf{#1}\hspace{0.5em}}

\vspace{-2mm}
\section{Introduction}
\label{sec:introduction}
\vspace{-1mm}
PPG-to-vital-sign reconstruction turns the photoplethysmogram of a consumer wearable device into clinical waveforms such as the ECG and respiration~\cite{nie_ppg_review_2024}, from which heart and respiratory rates are read. Long-horizon multivariate time-series forecasting supports planning in energy, weather-dependent operations, and traffic~\cite{Wu_intro_weather_2023, Martin_intro_energy_20101772}. The two tasks are studied by separate communities. Both, however, generate a target sequence from a condition sequence, and both hinge on one question: \emph{which part of the condition should each target position read?} Reconstruction must read the sample at the same position, and forecasting must read one seasonal period back. A conditioning rule that serves both is what a general time-series model needs to accept physiological waveforms as inputs, which current foundation models such as Chronos~\cite{ansari_chronos_2024}, trained on scalar series, do not.

Each community answers the question with a hard-coded mechanism. Reconstruction models~\cite{shome_rddm_2024, suzuki_penguin_2026} copy the conditioning content from the sample at the same position, an identity skip that is appropriate for synchronous signals but meaningless for forecasting, where the target begins after the condition ends. Forecasting models build in seasonal recurrence through decomposition~\cite{wu_autoformer_2021, zeng_dlinear_2023, nagashima_decompssm_2026}, so a target implicitly reads the condition one period back, fixed by the architecture and never measured. Neither mechanism transfers between tasks, neither fails gracefully when the correspondence is corrupted (motion in wrist PPG) or the data has no period (exchange rates), and neither is observable, so a wrong assumption surfaces only as test error.

\begin{figure*}[t]
\centering
\includegraphics[width=0.94\textwidth]{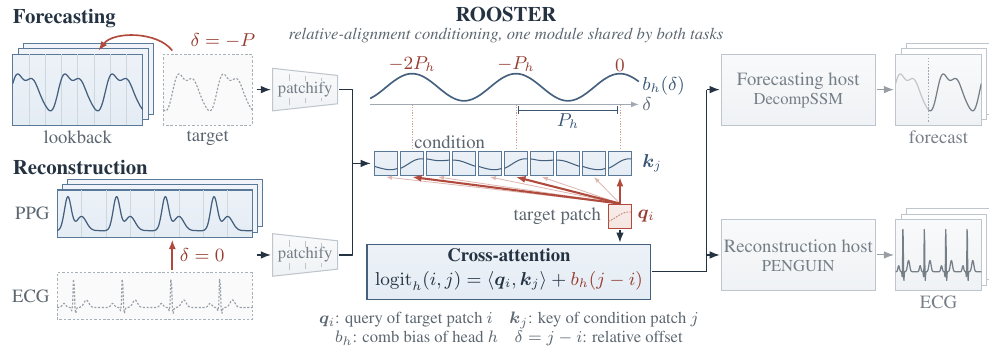}
\vspace{-4mm}
\caption{One learned correspondence for two tasks. Left: forecasting must read the lookback one period back ($\delta=-P$), and reconstruction must read the PPG at the same position ($\delta=0$). Middle: ROOSTER is cross-attention from target to condition patches whose logits carry a periodic-comb bias $b_h$ over the relative offset, so both correspondences are peaks of one learnable prior. Right: the module's output is added to each task's host, which is otherwise unchanged (gray). The learned $(\phi_h, P_h)$ are read out in Sec.~\ref{subsec:mechanism}.}
\label{fig:method}
\vspace{-3mm}
\end{figure*}

The two answers differ only in the value of a relative offset, $0$ for reconstruction and $-P$ for forecasting, so the offset should be \emph{learned}, by one module, and \emph{read out}. We introduce ROOSTER (Fig.~\ref{fig:method}), cross-attention from target to condition patches whose logits carry a learned periodic-comb bias over their offset, so the module can settle on the identity, a seasonal lag, or a mixture, and its learned period can be read out. We attach ROOSTER, one design with separately trained weights, to two task-specific host models (the networks it is inserted into) from our prior work, PENGUIN~\cite{suzuki_penguin_2026} for vital-sign reconstruction and DecompSSM~\cite{nagashima_decompssm_2026} for forecasting, and evaluate on the standard benchmarks and protocols of the corresponding papers. We make three contributions, and our code is publicly available.\footnote{\url{https://github.com/Neurogica/ROOSTER}}
\noindent
\begin{list}{$\bullet$}{\setlength{\leftmargin}{1em}\setlength{\itemsep}{0.25ex}\setlength{\topsep}{0.25ex}}
\raggedright
\item We treat the identity skip of reconstruction and the seasonal lag of forecasting as two settings of one learned relative offset, implement it as ROOSTER's periodic-comb bias in cross-attention, and read out which setting it found (one day on 15-minute data, the identity on reconstruction, and no period on the aperiodic exchange rates).
\item We evaluate ROOSTER on both tasks' standard benchmarks, where it outperforms the published baselines: on all four PPG reconstruction tasks under PENGUIN's metrics, and in horizon-averaged MSE on all four forecasting datasets.
\item We isolate the relative bias with an ablation study on both tasks, make the auxiliary objective diagnosable, and analyze the PPG-DaLiA error, ruling out five candidate repairs.
\end{list}

\vspace{-3mm}
\section{Related Work}
\label{sec:related_work}
\vspace{-2mm}

\parhead{PPG-based vital-sign reconstruction.}
PaPaGei~\cite{pillai_papagei_2025} predicts several vital signs from PPG, but as interval-level values that discard waveform morphology. Generative models reconstruct the waveform, RDDM~\cite{shome_rddm_2024} for PPG-to-ECG with diffusion and RespDiff~\cite{miao_respdiff_2025} for respiration, and PENGUIN~\cite{suzuki_penguin_2026}, our reconstruction host and baseline, unifies ECG, respiratory, and blood-pressure reconstruction under one conditional flow-matching~\cite{lipman_flow_matching_2023} model.

\parhead{Multivariate time-series forecasting.}
Modern forecasting models build in the classical decomposition principle~\cite{cleveland_tsf_90}: Autoformer~\cite{wu_autoformer_2021}, FEDformer~\cite{zhou_fedformer_2022}, DLinear~\cite{zeng_dlinear_2023}, and TimeMixer~\cite{wang_timemixer_2024} rely on a fixed moving-average split, while LaST~\cite{wang_last_2022} and CoST~\cite{woo_cost_2022} learn seasonal/trend structure in latent space. DecompSSM~\cite{nagashima_decompssm_2026} instead decomposes each variate with parallel input-adaptive S5~\cite{smith_s5_2023} branches plus cross-variable refinement. It is our forecasting host and baseline, and we change only its conditioning pathway (Sec.~\ref{subsec:module}).

\parhead{Relative position biases.}
Learned functions of the query--key offset are standard in language models, as free per-offset embeddings~\cite{shaw_relpos_2018, raffel_t5_2020} or fixed linear penalties~\cite{press_alibi_2022}. ROOSTER's bias is a \emph{periodic} parametric family whose learned period is itself a read-out (Sec.~\ref{subsec:ablations} compares the three). Autoformer's auto-correlation~\cite{wu_autoformer_2021} and TimesNet's FFT period detection~\cite{wu_timesnet_2023} also find lags from data, but they select discrete lags per input at inference, whereas ROOSTER learns one continuous period per head and reads it out.

\vspace{-3mm}
\section{Proposed Method}
\label{sec:methods}
\begin{table*}[t]
\centering
\caption{Quantitative comparison on the standard PPG-to-ECG and PPG-to-respiration reconstruction tasks. Baseline numbers are as reported in the PENGUIN paper~\cite{suzuki_penguin_2026}, and ours follow its metric protocol (Hamilton-method HR over 8\,s windows, 60\,s Fourier RR), subject-wise splits, seed 0. \textbf{Bold} and \underline{underlined} values denote the best and second-best performance. Parentheses give our margin to the best baseline.}
\label{tab:physio}
\setlength{\tabcolsep}{4.5pt}
\begin{tabular}{llcccccc}
\toprule
\multirow{2}{*}{Dataset} & \multirow{2}{*}{Metric} & \multicolumn{3}{c}{Specialist Model} & \multicolumn{2}{c}{Generalist Model} & \multirow{2}{*}{\textbf{ROOSTER}} \\
\cmidrule(lr){3-5}\cmidrule(lr){6-7}
 & & CycleGAN~\cite{aqajari_cyclegan_2021} & RDDM~\cite{shome_rddm_2024} & RespDiff~\cite{miao_respdiff_2025} & PaPaGei-S~\cite{pillai_papagei_2025} & PENGUIN~\cite{suzuki_penguin_2026} & \\
\midrule
\rowcolor{gray!12}\multicolumn{8}{l}{\emph{ECG Reconstruction}} \\
PPG-DaLiA~\cite{reiss_ppgdalia_2019} & \multirow{2}{*}{HR Error [bpm]} & 23.61 & 16.43 & 22.75 & 40.89 & \underline{15.64} & \textbf{9.77} \textcolor{penguingreen}{(-5.87)} \\
WildPPG~\cite{meier_wildppg_2024} &  & 23.21 & 16.02 & 20.57 & 55.42 & \underline{12.97} & \textbf{11.26} \textcolor{penguingreen}{(-1.71)} \\
\rowcolor{gray!12}\multicolumn{8}{l}{\emph{Respiratory Monitoring}} \\
BIDMC~\cite{pimentel_bidmc_2017} & \multirow{2}{*}{RR Error [bpm]} & 9.78 & 13.88 & 3.71 & 4.48 & \underline{2.98} & \textbf{1.65} \textcolor{penguingreen}{(-1.33)} \\
WESAD~\cite{schmidt_wesad_2018} &  & 11.93 & 10.12 & 5.12 & 5.84 & \underline{4.45} & \textbf{3.65} \textcolor{penguingreen}{(-0.80)} \\
\bottomrule
\end{tabular}
\end{table*}

\begin{table*}[t]
\centering
\caption{Multivariate time-series forecasting on four standard benchmarks (ECL, Weather, ETTm2, PEMS04): MSE/MAE in scaled space, lookback 96, averaged over horizons $\{96,192,336,720\}$ ($\{12,24,48,96\}$ on PEMS04). Baseline numbers are as reported in the DecompSSM paper~\cite{nagashima_decompssm_2026}, and ours are measured under the same protocol (Sec.~\ref{subsec:experimental_setup}, seed 0). Best in \textbf{bold}, second best \underline{underlined}.}
\label{tab:forecast}
\setlength{\tabcolsep}{3.2pt}
\footnotesize
\resizebox{\textwidth}{!}{%
\begin{tabular}{lcccccccccccccccc|cc}
\toprule
Dataset & \multicolumn{2}{c}{Autoformer~\cite{wu_autoformer_2021}} & \multicolumn{2}{c}{DLinear~\cite{zeng_dlinear_2023}} & \multicolumn{2}{c}{TimesNet~\cite{wu_timesnet_2023}} & \multicolumn{2}{c}{PatchTST~\cite{nie_patchtst_2023}} & \multicolumn{2}{c}{iTransformer~\cite{liu_itransformer_2024}} & \multicolumn{2}{c}{HDMixer~\cite{huang_hdmixer_2024}} & \multicolumn{2}{c}{PPDformer~\cite{wan_ppdformer_2025}} & \multicolumn{2}{c}{DecompSSM~\cite{nagashima_decompssm_2026}} & \multicolumn{2}{c}{\textbf{ROOSTER}} \\
 & MSE & MAE & MSE & MAE & MSE & MAE & MSE & MAE & MSE & MAE & MSE & MAE & MSE & MAE & MSE & MAE & MSE & MAE \\
\midrule
ECL & 0.227 & 0.338 & 0.231 & 0.323 & 0.216 & 0.311 & 0.216 & 0.304 & 0.176 & 0.268 & 0.194 & 0.294 & 0.168 & \underline{0.267} & \underline{0.167} & \textbf{0.261} & \textbf{0.166} & \textbf{0.261} \\
Weather & 0.338 & 0.382 & 0.266 & 0.316 & 0.258 & 0.285 & 0.258 & 0.280 & 0.261 & 0.282 & 0.255 & 0.282 & 0.246 & 0.276 & \underline{0.242} & \textbf{0.270} & \textbf{0.240} & \underline{0.271} \\
ETTm2 & 0.311 & 0.354 & 0.353 & 0.401 & 0.298 & 0.334 & 0.289 & 0.333 & 0.292 & 0.336 & 0.286 & 0.330 & 0.285 & 0.329 & \underline{0.280} & \textbf{0.320} & \textbf{0.279} & \underline{0.324} \\
PEMS04 & 0.610 & 0.589 & 0.295 & 0.388 & 0.129 & 0.241 & 0.195 & 0.307 & 0.120 & 0.232 & 0.163 & 0.281 & 0.104 & 0.213 & \underline{0.103} & \underline{0.212} & \textbf{0.100} & \textbf{0.211} \\
\bottomrule
\end{tabular}}
\end{table*}

\begin{table*}[t]
\centering
\caption{Ablation study with all other settings fixed. Left: reconstruction, the conditioning scheme is the only axis (rate MAE in bpm, matched-harness estimator, medoid readout, seed 0). Right: forecasting, the two proposed components removed from the full model (horizon-averaged MSE at the published protocol): the auxiliary quantile objective (Aux., Sec.~\ref{subsec:aux}) and the relative alignment module (Align., Sec.~\ref{subsec:module}). Best in \textbf{bold}, second best \underline{underlined}.}
\label{tab:cond_ablation}
\footnotesize
\setlength{\tabcolsep}{3pt}
\begin{minipage}[t]{0.645\textwidth}\centering
\begin{tabular}{lccccc}
\toprule
\multicolumn{6}{c}{Reconstruction (rate MAE, bpm)} \\
\cmidrule(lr){1-6}
Conditioning & BIDMC-ECG & Capno-ECG & BIDMC-RESP & Capno-RESP & DaLiA-ECG \\
\midrule
injection only & 7.28 & 3.44 & 2.19 & 4.78 & \textbf{7.98} \\
identity skip & \underline{3.06} & \underline{2.29} & \textbf{1.30} & \underline{3.44} & 15.63 \\
no relative bias & 6.90 & 3.49 & 1.58 & 4.69 & 14.02 \\
ROOSTER & \textbf{1.29} & \textbf{1.92} & \underline{1.56} & \textbf{3.22} & \underline{8.16} \\
\bottomrule
\end{tabular}
\end{minipage}\hfill
\begin{minipage}[t]{0.34\textwidth}\centering
\begin{tabular}{ccccc}
\toprule
\multicolumn{5}{c}{Forecasting (MSE, avg.\ over horizons)} \\
\cmidrule(lr){1-5}
Aux. & Align. & Weather & ETTm2 & PEMS04 \\
\midrule
-- & -- & 0.245 & 0.293 & 0.106 \\
$\checkmark$ & -- & \underline{0.241} & \underline{0.283} & 0.104 \\
-- & $\checkmark$ & 0.243 & \textbf{0.279} & \underline{0.101} \\
$\checkmark$ & $\checkmark$ & \textbf{0.240} & \textbf{0.279} & \textbf{0.100} \\
\bottomrule
\end{tabular}
\end{minipage}
\end{table*}

\vspace{-2mm}

\vspace{-1mm}
\subsection{Two tasks, one conditioning problem}
\vspace{-1mm}
Both tasks (Fig.~\ref{fig:method}) map an observed condition sequence to a target sequence. In reconstruction, the model observes a PPG segment and predicts a \emph{different} channel (ECG or respiration) over the \emph{same} timespan. In forecasting, it observes a lookback $\bm{X} \in \mathbb{R}^{T \times M}$ and predicts the next $H$ steps of the \emph{same} channels. Both sequences are split into non-overlapping patches of length $\ell$, and every conditioning mechanism must decide \emph{which condition patch target patch $i$ reads}: patch $i$ (offset $0$) for reconstruction, and patch $i-P$, one seasonal period back, for forecasting. Both are relative offsets, so we learn the offset.

\vspace{-2mm}
\subsection{ROOSTER: relative-alignment conditioning}
\label{subsec:module}
\vspace{-1mm}
ROOSTER is cross-attention from $N_t$ target-patch queries to $N_c$ condition-patch keys, with a bias that depends only on their relative offset $\delta = j - i$ (condition index $j$, target index $i$, so negative $\delta$ looks back). For head $h$,
\begin{equation}
\mathrm{logit}_h(i,j) = \frac{\bm{q}_i^\top \bm{k}_j}{\sqrt{d_k}} + b_h(j - i),
\end{equation}
where $\bm{q}_i$ and $\bm{k}_j$ are the query and key projections of target patch $i$ and condition patch $j$, and $d_k$ is the head dimension. Content similarity and positional prior are therefore additive. The bias is a \emph{periodic comb},
\begin{equation}
b_h(\delta) = -\kappa_h \left(1 - \cos \frac{2\pi (\delta - \phi_h)}{P_h}\right),
\label{eq:comb}
\end{equation}
where $\phi_h$, $P_h$, and $\kappa_h$ are the center, period, and sharpness of head $h$, all learned. The attention output passes through a zero-initialized projection into the host's conditioning pathway, so an untrained module leaves the host unchanged.

\parhead{Why a periodic comb.}
We want one bias form that expresses the correspondence of either task, stays learnable, and reports what it learned as a number. Reconstruction needs a single peak at offset $0$, where the target patch reads the PPG patch at the same position. Forecasting needs peaks at $-P, -2P, \dots$, because seasonal data repeat every $P$ steps. The cosine comb of Eq.~\eqref{eq:comb} covers both. When $P_h$ exceeds the window, only the peak at $\phi_h \approx 0$ remains and the comb is the single bump that reconstruction needs, and when $P_h$ matches the season it places one peak on every seasonal offset. The alternatives can match its accuracy (Sec.~\ref{subsec:ablations}) but not its read-out. A single bump centered at $0$ has no gradient at offset $-P$, so training can sharpen it but cannot move it, and a free per-offset table ties nothing and exposes no period. The comb ties all multiples through one parameter, has gradient at every offset, and exposes that parameter as a number. The $(\phi_h, P_h)$ of the head with the largest $\kappa_h$ (the sharpest head) are logged for every run, and comparing $P_h$ with the sampling interval tells whether the model found the identity, a day, or nothing (Sec.~\ref{subsec:mechanism}).

\vspace{-2mm}
\subsection{Host models}
\vspace{-1mm}
\parhead{Reconstruction host.}
A conditional flow-matching~\cite{lipman_flow_matching_2023} reconstruction model on PENGUIN's Flow-SSM backbone~\cite{suzuki_penguin_2026}: an S5 condition encoder over PPG patches, a velocity field over normalized patch values with per-layer conditioning injection, and sampled ensembles scored through a medoid readout (the member closest to all others). ROOSTER replaces PENGUIN's hard-coded skip from the same-position PPG patch.

\parhead{Forecasting host.}
Our re-implementation of DecompSSM~\cite{nagashima_decompssm_2026}, with $K$ parallel gated S5 branches, input-adaptive discretization, cross-variable refinement, DecompSSM's decomposition losses, and a summed component mixing head. The host folds each variate's lookback into one token, so we patch the normalized lookback per variate, attend with learned target-patch queries, and add the pooled result to the variate embedding.

\vspace{-2mm}
\subsection{A diagnosable auxiliary objective}
\label{subsec:aux}
\vspace{-1mm}
Alongside the primary MSE loss, the forecasting host carries an auxiliary pinball term on a monotone quantile head over the shared feature extractor (the trunk) that the point-forecast head also reads: for quantile levels $\tau \in \mathcal{T}$ and residual $u = y - \hat{y}_\tau$, $\mathcal{L}_{\mathrm{q}} = \sum_{\tau} \max(\tau u, (\tau - 1) u)$, added with a fixed weight. The term is a regularizer whose engagement can be measured. It restores the effective rank of the trunk (the number of feature directions in use) where that rank has collapsed and is inert where the trunk is healthy (Sec.~\ref{subsec:mechanism}), so trunk rank measured early predicts whether it helps.

\vspace{-3mm}
\section{Experiments}
\label{sec:experiments}
\vspace{-2mm}

\subsection{Setup}
\label{subsec:experimental_setup}

\parhead{Datasets and protocol.}
Reconstruction follows PENGUIN's metric protocol~\cite{suzuki_penguin_2026}: HR error from Hamilton-method beat detection over 8\,s windows, RR error from the Fourier dominant frequency over 60\,s windows, strictly subject-wise splits. Windows are 8\,s at 125\,Hz for ECG and 32--60\,s at 62.5\,Hz for respiration. Forecasting follows DecompSSM's setup~\cite{nagashima_decompssm_2026}: lookback 96, horizons 96--720 (12--96 on PEMS04), chronological splits, scaler fit on the training split only, MSE/MAE in scaled space over the full test split. ROOSTER is trained for 20k steps with Adam at seed 0, as in both prior papers, and separately for every dataset, horizon, and task. The two tasks share the module's form and objective, not its weights. Patches have length $\ell{=}8$, the reconstruction model scores 20-sample ensembles through the medoid readout, the forecasting host uses $K{=}3$ branches and auxiliary weight $1$, and $P_h$ is initialized geometrically across heads. Additional tasks (CapnoBase~\cite{karlen_capnobase_2013}, BIDMC-ECG~\cite{pimentel_bidmc_2017}) and the ETTm1, ETTh1, ETTh2, Exchange, and Solar datasets are evaluated under our own harness, which retrains every model, three seeds each on forecasting, with the same steps, samples, and scoring.

\vspace{-2mm}
\subsection{PPG-to-vital-sign reconstruction}
\vspace{-1mm}
\begin{table}[t]
\centering
\caption{Matched-harness control: rate MAE in bpm under one estimator and one medoid readout applied identically to both models (PENGUIN retrained from its original code), subject-wise splits, seed 0. Parentheses: fraction of windows with no resolvable rate, which trades against the MAE.}
\label{tab:physio_harness}
\setlength{\tabcolsep}{3.0pt}
\footnotesize
\resizebox{\linewidth}{!}{%
\begin{tabular}{lccccc}
\toprule
Model & BIDMC-ECG & Capno-ECG & BIDMC-RESP & Capno-RESP & DaLiA-ECG \\
\midrule
PENGUIN (retrained) & 4.76 (0.09) & 3.17 (0.09) & \textbf{1.21} (0.00) & 3.39 (0.00) & 8.37 (0.00) \\
ROOSTER & \textbf{1.29} (0.01) & \textbf{1.92} (0.01) & 1.56 (0.00) & \textbf{3.22} (0.00) & \textbf{8.16} (0.02) \\
\bottomrule
\end{tabular}}
\end{table}

Table~\ref{tab:physio} evaluates ROOSTER on the standard PPG-to-vital-sign reconstruction tasks under PENGUIN's metric protocol, on our fixed subject-wise split, against the numbers that the published baselines~\cite{aqajari_cyclegan_2021, shome_rddm_2024, miao_respdiff_2025, pillai_papagei_2025} report on their own splits: ECG reconstruction on PPG-DaLiA~\cite{reiss_ppgdalia_2019} and WildPPG~\cite{meier_wildppg_2024}, and respiratory reconstruction on BIDMC and WESAD~\cite{schmidt_wesad_2018}. PENGUIN's two ABP rows were omitted because ABP requires amplitude calibration in mmHg, which our normalized reconstruction does not provide. ROOSTER was best on all four tasks, and in every case the strongest baseline was PENGUIN (9.77 vs.\ 15.64 bpm HR error on PPG-DaLiA, 1.65 vs.\ 2.98 bpm RR error on BIDMC).

Table~\ref{tab:physio_harness} adds a controlled comparison: PENGUIN retrained from its original code and scored under our harness with the same rate estimator and medoid readout, plus the fraction of unresolved windows. ROOSTER outperformed PENGUIN on 4 of 5 tasks, with PPG-DaLiA close to a tie (8.16 vs.\ 8.37). On the two clinical ECG tasks, PENGUIN returned no rate on roughly nine times as many windows and was still outperformed, so the reported gap understates the actual difference.

\vspace{-2mm}
\subsection{Forecasting}
\vspace{-1mm}
Table~\ref{tab:forecast} evaluates ROOSTER on the four standard forecasting benchmarks used by DecompSSM, against the same baselines~\cite{wu_autoformer_2021, zeng_dlinear_2023, wu_timesnet_2023, nie_patchtst_2023, liu_itransformer_2024, huang_hdmixer_2024, wan_ppdformer_2025}. Averaged over horizons, ROOSTER had the best MSE on all four datasets, and the best MAE on PEMS04 and ECL (tied on the latter).
Under one harness at matched budget over six datasets (the four ETT sets, Weather, Exchange) and four horizons, with three seeds per model, ROOSTER had lower mean MSE than retrained DecompSSM on 20 of 24 dataset--horizon settings. Eight margins exceeded the sum of both models' standard deviations, all in our favor, and none of the four opposite ones did. The margins were largest on ETTh2 (7--17\% lower MSE at every horizon, all beyond the seed spread), 3--5\% on ETTm2 and 1--5\% on ETTm1, and within seed noise on ETTh1 and Weather. The only dataset on which it was worse, Exchange (by 5--11\% at three of four horizons), was the one where the module found no period, as reported in Sec.~\ref{subsec:mechanism}.

\vspace{-2mm}
\subsection{Analysis of the learned module}
\label{subsec:mechanism}
\vspace{-1mm}
\parhead{The learned offset is task-appropriate and physically plausible.}
On 15-minute data, one patch is 2 hours and one day is exactly 12 patches. The sharpest head learned $P{=}13.0$ on ETTm1 and $11.1$ on ETTm2, both within 10\% of one day. On hourly ETTh1/ETTh2, where a day spans only three 8-hour patches, the learned period (13.7/15.4) exceeded the 12-patch lookback, so the comb degenerated to one broad peak. On Exchange, the one dataset without seasonal structure, the learned $P{=}1.7$ corresponds to no period, and this is also where the module degraded accuracy. On reconstruction the module received no such prior, yet on the three ECG tasks with clean PPG (BIDMC, CapnoBase, WildPPG) the sharpest head's center $\phi_h$ settled within 0.1 patches of zero, under one sample. The module rediscovered from data the same-position correspondence that PENGUIN hard-codes as a skip, and then outperformed that skip (Sec.~\ref{subsec:ablations}). On respiration the center stayed within 1.4 patches (0.2\,s). On PPG-DaLiA, where motion corrupts the correspondence, it drifted to 5.9 patches, the case that Sec.~\ref{subsec:error} analyzes.

\parhead{The auxiliary term repairs rank collapse where it occurs.}
We measured the effective rank of the shared trunk (of 128) with and without the auxiliary term at matched seed: Solar $1.93 \to 2.56$ (MSE effect unstable), ETTm2 $3.57 \to 6.26$ (MSE $-6.4$\%), and ETTm1 $37.73 \to 30.01$ (MSE $+1.0$\%). These values come from checkpoints of the harness runs (Sec.~\ref{subsec:experimental_setup}), not from Tables~\ref{tab:forecast} or~\ref{tab:cond_ablation}. The term restored rank and lowered error where the trunk had collapsed, and where it was healthy (ETTm1) it did not help.

\vspace{-2mm}
\subsection{Ablation study of the conditioning scheme on both tasks}
\label{subsec:ablations}
\vspace{-1mm}
In Table~\ref{tab:cond_ablation} (our variants only), the conditioning scheme is the only axis on reconstruction (left). Removing only the relative bias collapsed BIDMC-ECG from 1.29 to 6.90\,bpm and PPG-DaLiA from 8.16 to 14.02, so the bias, not content matching, carried the alignment. ROOSTER also outperformed the identity skip it generalizes on all three ECG tasks (BIDMC-ECG 1.29 vs.\ 3.06). On PPG-DaLiA, where the center drifted (Sec.~\ref{subsec:mechanism}), injection alone was marginally better (7.98 vs.\ 8.16). On forecasting (right), both components were switched off in all four combinations. Removing both cost 2\% on Weather, 5\% on ETTm2, and 6\% on PEMS04, and the full model was best in every column. The two were complementary rather than additive. On ETTm2 the module alone recovered the gap (0.279), on PEMS04 most of it (0.101), and on Weather the auxiliary term did (0.241), consistent with Sec.~\ref{subsec:mechanism}. On reconstruction, the \emph{form} of the bias mattered far less than its presence. A free per-offset table (as in~\cite{shaw_relpos_2018, raffel_t5_2020}) or a single aperiodic bump (a local penalty, as in ALiBi~\cite{press_alibi_2022}) changed $H{=}96$ MSE by at most 0.003 (three seeds) on five of six harness datasets, and only on ETTh2 was the comb clearly best (0.303 vs.\ 0.309 and 0.321).

\vspace{-2mm}
\subsection{Error analysis: PPG-DaLiA}
\label{subsec:error}
\vspace{-1mm}
PPG-DaLiA and WildPPG, the two wrist recordings under free-living motion, have errors an order of magnitude above the clinical ECG tasks (9.77 and 11.26 bpm). DaLiA also provides activity labels and is the one task on which the learned center drifted (Sec.~\ref{subsec:mechanism}), so we analyzed it. The residual error was a signed, activity-structured bias: $-20.9$ bpm on stairs (true rate 117), $+6.7$ while sitting (56), both regressing toward the mean training heart rate of 78 bpm. An oracle choice among the ensemble members per window would score 1.63 bpm against the medoid's 8.16. Five candidate repairs (spectral rate anchoring, a Viterbi readout across windows, amplified conditioning at sampling, accelerometer covariates, and notching the step cadence) left the bias unchanged, which suggests that at this window length motion-corrupted PPG retains little rate information that a PPG-only readout could recover.

\vspace{-2mm}
\subsection{Limitations}
\vspace{-1mm}
The learned prior drifts when the correspondence itself is corrupted, as under motion (Sec.~\ref{subsec:mechanism}), and the module cannot detect this at test time. Our DaLiA numbers are PPG-only, not comparable with accelerometer-assisted results. The harness comparisons used three seeds, but Tables~\ref{tab:physio} and~\ref{tab:forecast} report single-seed runs at the published protocols. We have not investigated BIDMC-RESP, the one task where PENGUIN was better under the harness. The period read-out needs a lookback that spans several periods. On hourly ETTh data, with three patches per day, it did not resolve the day (Sec.~\ref{subsec:mechanism}).

\vspace{-3mm}
\section{Conclusion}
\vspace{-2mm}
We presented ROOSTER, one conditioning module for both PPG-to-vital-sign reconstruction and long-horizon forecasting, built on a learned periodic-comb bias over the target--condition offset. ROOSTER learned the identity alignment on reconstruction and the daily period on sub-hourly forecasting data, improved both task-specific hosts under matched budgets (4 of 5 tasks against PENGUIN, 20 of 24 dataset--horizon settings against DecompSSM), and its learned period indicates whether the data contains a usable correspondence. Because ROOSTER only requires patch sequences related by an offset, it could extend to other conditional sequence tasks and to time-series foundation models. In future work, we plan to make the conditioning robust when the correspondence itself is corrupted, for example under motion in wrist-worn PPG.

\bibliographystyle{IEEEbib}
\bibliography{strings,refs}

\end{document}